\documentclass[journal,letterpaper]{IEEEtran}

\usepackage[noadjust]{cite}

\usepackage[utf8]{inputenc}
\usepackage[T1]{fontenc}   
\usepackage{hyperref}       
\usepackage{url}           
\usepackage{booktabs}       
\usepackage{amsfonts}       
\usepackage{nicefrac}       
\usepackage{microtype}

\usepackage{amssymb,bm}
\usepackage{amsthm,algorithm,algorithmic,bbm,mathtools,amsmath}

\newcommand{\bs}[1]{\boldsymbol{#1}}
\newcommand{\vt}{{\bs{\theta}}}

\DeclareMathOperator*{\argmin}{arg\,min}
\DeclareMathOperator*{\argmax}{arg\,max}
\newcommand{\Tau}{\mathcal{T}}
\newcommand{\Wau}{\mathcal{W}}
\newcommand{\Hau}{\mathcal{H}}
\newcommand{\Nau}{\mathcal{N}}

\newcommand{\lt}[1][t]{l_{#1}}
\newcommand{\ft}[1][t]{f_{#1}}

\newtheorem{theorem}{Theorem}
\newtheorem{lemma}[]{Lemma}
\newtheorem{corollary}[]{Corollary}
\newtheorem{remark}[]{Remark}
\newtheorem{definition}[]{Definition}

\newtheorem{assumption}[]{Assumption}
\newtheorem{example}[]{Example}

\title{Near-Linear Time Algorithm with Near-Logarithmic Regret Per Switch for Mixable/Exp-Concave Losses}

\author{
	\IEEEauthorblockN{Kaan Gokcesu}, \IEEEauthorblockN{Hakan Gokcesu}
  }

\begin{document}

\maketitle

\begin{abstract}
	We investigate the problem of online learning, which has gained significant attention in recent years due to its applicability in a wide range of fields from machine learning to game theory. Specifically, we study the online optimization of mixable loss functions with logarithmic static regret in a dynamic environment. The best dynamic estimation sequence that we compete against is selected in hindsight with full observation of the loss functions and is allowed to select different optimal estimations in different time intervals (segments). We propose an online mixture framework that uses these static solvers as the base algorithm. We show that with the suitable selection of hyper-expert creations and weighting strategies, we can achieve logarithmic and squared logarithmic regret per switch in quadratic and linearithmic computational complexity, respectively. For the first time in literature, we show that it is also possible to achieve near-logarithmic regret per switch with sub-polynomial complexity per time.	Our results are guaranteed to hold in a strong deterministic sense in an individual sequence manner.
\end{abstract}

\section{Introduction}
In machine learning literature \cite{jordan2015machine,mohri2018foundations}, the problems of online learning, prediction and estimation \cite{shalev2011online,cesa_book, poor_book} are heavily investigated because of their applications in a wide variety of fields including decision theory \cite{tnnls4}, game theory \cite{tnnls1,chang}, control theory \cite{tnnls3}, optimization \cite{zinkevich,hazan}, density estimation or source coding \cite{gokcesu2017density,willems,coding1,coding2}, anomaly detection \cite{gokcesu2018anomaly}, adversarial bandits \cite{cesa-bianchi,neyshabouri2018asymptotically,vural2019minimax,cesa2007,gokcesu2018adversarial}, prediction \cite{singer,singer2} and signal processing \cite{gokcesu2018adaptive,ozkan,signal1}.

\subsection{Online Learning}
In online learning, at each time $t$ of the decision process, we make a prediction $\vt_t$ and incur a corresponding loss \cite{cesa_book}. Since the loss functions at each round of the decision process may differ, they are generally denoted as a time varying process, i.e., the loss incurred is $l_t(\vt_t)$. As opposed to the batch learning setting, we need to produce these decision $\vt_t$ sequentially, where each prediction $\vt_t$ can only depend on our past predictions $\{\vt_\tau\}_{\tau=1}^{t-1}$ and their corresponding observations (i.e., the predictions are made before observing the resulting loss). Depending on the problem setting, we may either have access to the whole description of the losses $l_t(\cdot)$, only their evaluations $l_t(\vt_t)$ or some other auxiliary information such as the gradient $\nabla l_t$ or the Hessian $\nabla^2l_t$. The goal is to create the predictions $\vt_t$ in such a way that we minimize the cumulative loss incurred, i.e., $\sum_{t=1}^{T}l_t(\vt_t)$. This phenomena can be better understood by the online portfolio management problem \cite{li2014}. In the online portfolio management, the equity at hand is distributed among several stocks at time $t$. However, we cannot see the reward (or the loss) of this action until we see the price changes in the stock market. Hence, the decisions need to be made sequentially in an online manner. 

\subsection{Mixable Losses}\label{sec:loss}
There is a myriad of different losses utilized in the literature and they are generally classified according to some nice properties they possess, which helps the learning process. For example, because of its general applicability to different problem scenarios and relatively straightforward solutions, convex losses has been extremely popular in the online learning literature \cite{boyd2004}. However, this applicability, which stems from its comparatively milder conditions, may result in slow learning. Some stricter conditions have been studied starting with the notion of strong-convexity for faster convergence in online learning literature \cite{strong_convex}. However, while these losses result in a much faster learning, their conditions can become too strict and limit their applicability \cite{cesa_book,poor_book}. To this end, the conditions of the strong-convexity have been loosened by the notion of mixability \cite{vovk}. When a loss $l(\cdot)$ is $\alpha$-mixable, they conform to the following definition.
\begin{definition}\label{def:mixable}
	$l(\bs{\theta})$ is $\alpha$-mixable in $\bs{\theta}$ if we have
	\begin{align*}
		e^{-\alpha l(\hat{\bs{\theta}})}\geq\sum_{i}{P}_ie^{-\alpha l(\bs{\theta}_i)}.
	\end{align*} 
	where $\{P_i\}_{i=1}^n$ is some probability distribution and
	\begin{align*}
		\hat{\bs{\theta}}=F(\{\bs{\theta}_i,P_i\}_{i=1}^n),
	\end{align*}
	for a surrogate function $F(\cdot)$.
\end{definition}
The surrogate function or mixture rule $F(\cdot)$ in \autoref{def:mixable} is specific for different losses. Few examples are as follows.
\begin{example}
	The square loss function, where $\vt\in[-1,1]$, and $l(\vt)=(\vt-x)^2$ (for some $x$) is mixable with $\alpha=\frac{1}{2}$ \cite{vovk, haussler1998}, where 
	\begin{align*}
		&\hat{\vt}=\frac{1}{2}\sum_{q=-1}^{1}q\log\left(\sum_{i=1}^{n}P_ie^{-\frac{1}{2}(\vt_i-q)^2}\right)
	\end{align*}
	is the aggregate estimation.
\end{example}

\begin{example}\label{ex:den}
	Density estimation (or source coding) under log-loss \cite{willems} is $1$-mixable with the estimator
	\begin{align*}
		\hat{\vt}=\sum_{i}P_i\vt_i,
	\end{align*}
	where the mixability inequality holds with equality.
\end{example}
\begin{example}
	Exp-concave loss functions \cite{koren2013} are mixable, where $\lambda$-exp-concave loss is $\lambda$-mixable with the mean estimator as in \autoref{ex:den} since
	\begin{align*}
		e^{-\lambda l(\hat{\vt},x)}\geq\sum_{i}P_ie^{-\lambda l(\vt_i,x)},
	\end{align*}
	from the exp-concavity.
\end{example}

\subsection{Notion of Regret and Oracle}
In the problem of online learning, the loss functions can be arbitrarily high and achieving small losses may be quite challenging if not impossible. To this end, it has been the norm to aim for a loss as much as a viable competition \cite{cesa_book}. Although infeasible, we can achieve the best performance when we make the decisions with full knowledge of the losses in advance. To this end, the challenge is to get as close as possible to that best performance, which is called the oracle loss. Hence, the notion of regret has been used in a competitive framework. The notion of regret is defined as the difference between the losses incurred by an algorithm and the losses incurred by a competition oracle. Against the oracle selections $\{\vt^*_t\}_{t=1}^T$, the cumulative regret up to time $T$ is given by
\begin{align}
	R_T\triangleq\sum_{t=1}^T r_t=\sum_{t=1}^T l_t(\vt_t)-\sum_{t=1}^T l_t(\vt^*_t).\label{eq:Rt}
\end{align}

Note that, we can not achieve small regret against arbitrary oracle selections and some form of regularity needs to be applied. One example is the fixed oracle (static), where we compete against a fixed estimation $\vt^*$, i.e., $\vt^*_t=\vt^*$ for all $t$. The literature on these static learners is abundant and its applications are numerous \cite{cesa_book}. In our work, we are specifically interested in the static learners with logarithmic regret bounds, i.e., $R_T=O(\log(T))$ such as \cite{vovk,hazan}. A harder competition would be against dynamically switching oracle estimations, where $\vt^*_t$ can change arbitrarily during a time horizon $T$. In this setting, the static learners are prone to stagnation and different strategies are utilized \cite{comp2,freund1997,lehrer2003,blum2007,hazan2009}. 

\subsection{Competing Against Dynamically Switching Estimations}
Although static learners are able to perform satisfactorily when competing against fixed oracles, they can under-perform when competing against dynamically changing oracle predictions. In some problem settings, making a static learner adaptive is pretty straightforward in lieu of parameter tuning such as online convex optimization \cite{zinkevich}. However, it is not the case for other problem scenarios \cite{cesa_book,poor_book}. For this purpose, there are various approaches in literature.

The sleeping experts in \cite{chernov2009} starts by creating a pool of hyper-experts, each of which mimics the learner's behavior (final output) during the first $t-1$ rounds (i.e., they are specialist experts \cite{freund1997using} that abstains from prediction up to time $t$). Each expert provides predictions from time $t$ onward and combined for adaptivity to create the final output estimations. These sleeping experts can be generalized to use different time selection functions as well \cite{blum2007}. 
The restarting experts in \cite{hazan2009} utilizes a main base algorithm and start a copy of it each time $t$. Then, by aggregating the predictions of these copies, the final output predictions are produced and adaptivity is achieved. It has been shown that, both of these approaches reduce to the Fixed Share algorithm \cite{adamskiy} with variable switching rates. The recursive experts in \cite{gokcesu2020recursive} creates the prediction outputs recursively by mixing a static expert, i.e., a single run of the base algorithm, with a dynamic expert, i.e., subsequent reruns of the base algorithm, recursively.

Therefore, the final predictions are constructed by selecting a set of time intervals which run a base algorithm and aggregating their individual predictions. The design of the time intervals requires care. While the set of intervals should be large enough to cover the whole time horizon sufficiently, they should not number too many, which may substantially increase the computational complexity. In the works of \cite{gyorgy2012,zhang2018}, the authors developed ways to construct intervals by trading effectiveness with efficiency. 
The geometric interval selections (generally powers of $2$) has become popular for efficient coverage of the time horizon \cite{daniely,zhang2019,zhang2020}.
Moreover, aggregation itself is a heavily investigated topic and there are techniques like Exponential Weights Algorithm (EWA) and Multiplicative Weights Algorithm (MWA) \cite{gokcesu2020generalized,daniely,littlestone1994,auerExp}.

\subsection{Contributions and Organization}
For the mixable losses, the optimal achievable regret stems from knowing the time instances the competition changes and individually learning them. With universal prediction and mixture techniques, it is possible to achieve this optimal regret bounds up to a finite multiplicative redundancy. Even so, these approaches have linear per time (or quadratic) computational complexities, which is inefficient. To this end, efficient time interval selections and their merger have been studied such as geometric coverage techniques. In spite of their efficient logarithmic per time (linearithmic) computational complexities, their regret bounds may become sub-optimal when the static problem has logarithmic regret. Thus, our goal is to combine the best of both worlds and design algorithms that are as efficient as possible whilst having near optimal regret. While the work in \cite{gokcesu2021mixable} specifically studies this problem of dynamic regret under mixable losses, their results are most meaningful when the base algorithm regret is sub-logarithmic or super-logarithmic. In this work, we bridge this gap.

\subsubsection{Problem Description}
In \autoref{sec:pre}, we provide some useful preliminaries; which include the mathematical formulations of the static and dynamic problems together with a general static solver, which will be utilized as a base algorithm.
\subsubsection{Algorithmic Framework}
In \autoref{sec:alg}, we provide the algorithmic framework, which incorporates a hyper-expert scheme (that runs the base algorithm) and a weighting scheme (their aggregation method). We also provide some important definitions for the regret analysis.
\subsubsection{Algorithm Design}
In \autoref{sec:des}, we provide generalized designs for the hyper expert scheme and the weighting scheme to compete against dynamically changing estimations. We also analyze their computational complexity and regret.
\subsubsection{Inefficient Optimal}
In \autoref{sec:lin}, we provide a quadratic time algorithm that can achieve optimal regret bounds much like \cite{willems,coding1,gokcesu2021mixable}.
\subsubsection{Efficient Sub-optimal}
In \autoref{sec:log}, we provide a linearithmic time algorithm that can achieve sub-optimal regret bounds as in \cite{hazan2009,gokcesu2020recursive,gokcesu2021mixable}
\subsubsection{Efficient Near-Optimal}
In \autoref{sec:opt}, for the first time in literature, we provide a near linear time algorithm that can achieve near-optimal regret bounds.
\subsubsection{Conclusion}
In \autoref{sec:con}, we finish with some concluding remarks.

\section{Problem Description}\label{sec:pre}
In this section, we mathematically describe the problems when we compete against a static (fixed) and a dynamic (changing) oracle estimations similar to \cite{gokcesu2020recursive,gokcesu2021mixable}. We also introduce a generic base algorithm as a building block.
\subsection{Logarithmic Regret Base Algorithm for the Static Problem}\label{sec:staticprob}
We start by first tackling the static problem setting. Let $\vt_t$ be some estimation we produce at the time $t$, and let
\begin{align}
	\vt_1^T\triangleq\{\vt_1,\ldots,\vt_T\}\label{x1T}
\end{align}
be our estimations for the whole time horizon, i.e., from $t=1$ to $T$. Let $\lt(\vt_t)$ be a finite loss we incur from our estimation $\vt_t$ and the cumulative loss for the time horizon be given by
\begin{align}
	L_T(\vt_1^T)\triangleq &\sum_{t=1}^{T}\lt(\vt_t).\label{Lt}
\end{align} 
Let $\vt^*$ be the best fixed estimation chosen in hindsight, i.e.,
\begin{align}
	\vt^*=\argmin_{\vt}\sum_{t=1}^{T}\lt(\vt)\label{x*},
\end{align}
and its cumulative loss in the time horizon be given by
\begin{align}
	L_T(\vt^*)\triangleq &\sum_{t=1}^{T}\lt(\vt^*).\label{Lx*}
\end{align} 
The regret of the static problem is given by the difference of these cumulative losses in \eqref{Lt} and \eqref{Lx*}, i.e.,
\begin{align}
	R(T,\vt^*)\triangleq L_T(\vt_1^T)-L_T(\vt^*).\label{RTx*}
\end{align} 

Let us have an algorithm (building block) that can solve the static problem. The base algorithm creates its estimation $\vt_t$ and incurs the corresponding loss $\lt(\vt_t)$ at time $t$. After that, it sequentially updates its estimation such that $\vt_{t+1}$ is a function of the past estimations $\{\vt_{t},\vt_{t-1},\ldots,\vt_1\}$, losses $\{\lt(\vt_t),\lt[t-1](\vt_{t-1}),\ldots,\lt[1](\vt_1)\}$ and possibly some auxiliary parameters $\{\lambda_t,\ldots,\lambda_1\}$. The update is formally given by
\begin{align}
	\vt_{t+1}=f(\vt_t,\ldots,\vt_1;\lt[t](\vt_t),\ldots,\lt[1](\vt_1);\lambda_t,\ldots,\lambda_1).\label{xt}
\end{align}
Note that, the update function $f(\cdot)$ depends on the structure of the base algorithm and the underlying problem setting (i.e., can vary in distinct applications). In sequential updates, the next estimation $\vt_{t+1}$ depends only on the current estimation and the algorithm's current state, i.e., it is given by
\begin{align}
	\vt_{t+1}=\ft[t](\vt_t),
\end{align}
for some $\ft(\cdot)$, which models the evolving algorithmic state (including the observations such as the loss). A summary is given in \autoref{alg:base}. We make the following assumption about the algorithm in \autoref{alg:base}.
\begin{assumption} \label{ass:RB}
	For the problem in \eqref{RTx*}, \autoref{alg:base} has logarithmic regret bounds, i.e.,
	\begin{align}
		R_{Base}(T)= O(\log(T)),\nonumber
	\end{align}
	for any $T$, where $O(\cdot)$ denotes the big-O notation, i.e., asymptotically, $R_{Base}(T)\leq K\log(T)$ for some finite $K$.
\end{assumption}

As opposed to the polynomial \cite{gokcesu2020recursive} or generic \cite{gokcesu2021mixable} regret, we specifically study logarithmic regret base algorithms.

\begin{algorithm}[!t]
	\caption{The Base Algorithm}\label{alg:base}
	{\begin{algorithmic}[1]
			\STATE Initialize internal parameters, $\vt_1$
			\FOR {$t=1,2,\ldots$}
			\STATE Observe $\lambda_t$, $\lt(\vt_t)$
			\STATE Determine $\ft[t](\cdot)$
			\STATE Update $\vt_{t+1}=\ft(\vt_t)$
			\ENDFOR
	\end{algorithmic}}
\end{algorithm}
\subsection{The Dynamic Problem and Its Achievable Regret Bound}\label{sec:dynamicprob}
Here, we explain the dynamic version of the static problem in \eqref{RTx*}. While our estimations $\vt_1^T$ in \eqref{x1T} and their cumulative loss $L_T(\vt_1^T)$ in \eqref{Lt} are as before, the competition differs. Instead of a fixed estimation, let us compete against a dynamically changing estimation. Let $S$ denote the number of times the competing estimations $\vt^*$ change throughout the time horizon $T$. Let us define $\vt_s^*$ for $s\in\{1,\ldots,S\}$ as the competing estimations for individual distinct $S$ time segments that cover the time horizon $T$. Let $t_s$ denote the length of the $s^{th}$ segment and $T_s$ be the sum of $t_{s'}$ from $s'=1$ to $s$ (i.e., $s^{th}$ segment ends on $T_s$). The expression in \eqref{Lx*} changes as
\begin{align}
	L_T(\{\vt_s^*,t_s\}_{s=1}^S)=\sum_{s=1}^S\sum_{T_{s-1}+1}^{T_s}l_t(\vt_s^*),
\end{align}
where $T_s$ is $0$ for $s=0$. Thus, our regret definition in \eqref{RTx*} changes to
\begin{align}
	R(T,\{\vt_s^*,t_s\}_{s=1}^S)\triangleq L_T(\vt_1^T)-L_T(\{\vt_s^*,t_s\}_{s=1}^S).\label{RTx*c}
\end{align}

Although, \autoref{alg:base} is able to compete against an optimal fixed parameter (the static problem), it may fail to compete against a dynamically changing estimation throughout the time horizon (the dynamic problem). For this problem, if we know of the time instances the competing parameter sequence changes (i.e., the exact times where we compete against distinct estimations $\vt_s^*$ for $s\in\{1,\ldots,S\}$), we can use \autoref{alg:base} as it is and restart it at the beginning of these time segments.
When the regret of the base algorithm for the static problem is as in \autoref{ass:RB}. We have the following for the dynamic problem.
\begin{remark}\label{thm:opt}
	If we know the time instances $T_s$ when the competition changes, we can restart \autoref{alg:base} in these times to get the regret bound
	\begin{align}
		R_{Base}(T,S)=O\left(S\log\left(\frac{T}{S}\right)\right),\nonumber
	\end{align} 
	which, in general, is the best regret bound achievable with \autoref{alg:base} for the dynamic problem in \eqref{RTx*c}.
\end{remark}
 
Note that, irrespective of whether the base algorithm itself is optimal or not, the result in \autoref{thm:opt} is the optimal regret achievable by using \autoref{alg:base}. Furthermore, the number of changes $S$ should be at least sub-linear (i.e., $o(T)$) for possibly viable learning, i.e., sub-linear regret bounds.

In some specific problem settings, it is straightforward to make a base algorithm adaptive to changes (such as the convex optimization \cite{zinkevich}). However, for general learning systems, it is not trivial \cite{cesa_book, poor_book,gokcesu2020recursive,gokcesu2021mixable}. In the next section, we will explain in detail how to incorporate this adaptivity.

\section{Algorithmic Framework}\label{sec:alg}
\subsection{Mixture of Hyper-Experts}\label{sec:mix}
The framework starts similarly to \cite{gokcesu2021mixable}, where we use a hyper-expert scheme to create a set of experts that use the base algorithm in \autoref{alg:base}.
\begin{definition}\label{def:Hau}
	The experts $i$ are created from the base algorithm according to an hyper-expert scheme $\Hau$ such that
	\begin{align}
		\Hau: H(i)=\{\bs{\lambda}_i\}, &&\forall i,t,
	\end{align} 
	where $\bs{\lambda}_i$ collectively defines the necessary information about how to utilize the base algorithm for the $i^{th}$ expert (e.g., start and finish times of the individual runs of the base algorithm). 
\end{definition}
At any time $t$, let us have a set $\Nau_t$ of parallel running hyper-experts, where each hyper-expert $i\in\Nau_t$ provides us with its parameter estimate $\bs{\theta}_{i,t}$. To create our final estimate $\bs{\theta}_t$ at time $t$, we combine their estimates $\bs{\theta}_{i,t}$ with some mixture probability distribution $P_{i,t}$ (i.e., $\sum_{i\in\Nau_t}^{} P_{i,t}=1$) using the surrogate function $F(\cdot)$. Hence, $\vt_t$ is given by
\begin{align}
	\vt_t=F(\{\bs{\theta}_{i,t},P_{i,t}\}_{i\in\Nau_t}^{}).\label{thetahat}
\end{align}
$P_{i,t}$ are given by the normalization of the weights $\widetilde{P}_{i,t}$, i.e.,
\begin{align}
	P_{i,t} &= \frac{\widetilde{P}_{i,t}}{\sum_{j\in\Nau_t}^{} \widetilde{P}_{j,t}}, && \text{for $i\in\Nau_t$}.\label{Pit}
\end{align}
Each hyper-expert weight $\widetilde{P}_{i,t}$ is designed to be dependent on the past performance of that expert. Furthermore, we also mix these weights with each other to compete against a dynamically changing estimation sequence. The performance weights are recursively calculated with a telescoping rule as
\begin{align}
	\widetilde{P}_{j,t} &= \sum_{i\in\Nau_{t-1}}^{} \widetilde{P}_{i,t-1} e^{-\alpha l_{t-1}(\bs{\theta}_{i,t-1})} \tau_t(i,j)\label{PitRec}
\end{align}
where $\tau_t(i,j)$ is the transition from the $i^{th}$ to $j^{th}$ hyper-expert.

\begin{definition}\label{def:Tau}
	The transition weights $\tau_t(i,j)$ for all $(i,j,t)$ are collectively defined as a weighting scheme $\Tau$ such that
	\begin{align}
		\Tau:\enspace \tau_{t}(i,j), &&\forall i,j,t,
	\end{align}
	where $\tau_t(i,j)$ is nonnegative for all $(i,j,t)$ and upper-bounded by a probability distribution, i.e.,
	\begin{align}
		\sum_{j}^{}\tau_t(i,j)\leq1, &&\forall j, t.
	\end{align}
\end{definition}

\begin{lemma}\label{thm:mixTrans}
	When the mixture uses the weighting scheme $\Tau$, which defines the transition weights $\tau_t(\cdot,\cdot), \enspace\forall t$; we have the following upper bound on our losses in terms of the losses of a sequence of hyper-expert selections $\{I_1,I_2,\ldots,I_T\}$ from the expert sets $\{\Nau_1,\Nau_2,\ldots,\Nau_T\}$ (from the scheme $\Hau$)
	\begin{align}
		\sum_{t=1}^T l_t({\bs{\theta}}_t) \leq\sum_{t=1}^T l_t(\bs{\theta}_{I_t,t})+\frac{1}{\alpha}\Wau_\Tau\left(\{I_t\}_{t=1}^T\right).\nonumber
	\end{align}
	where $\Wau_\Tau\left(\{I_t\}_{t=1}^T\right)\triangleq-\log\left(\prod_{t=1}^T \tau_t(I_{t-1},I_t)\right)$, $I_t\in\Nau_t$, $\tau_1(I_0,I_1)\leq P_{I_1,1}$ and each $l_t(\cdot)$ is $\alpha$-mixable.
	\begin{proof}
		The proof is in \cite{gokcesu2021mixable}.
	\end{proof}
\end{lemma}

\subsection{Important Definitions}
In an $S$ segment competition during a time horizon $T$, we compete against a dynamically changing oracle predictions $\{\vt_s^*\}_{s=1}^S$ for the duration of the time lengths $\{t_s\}_{s=1}^S$. The result of \autoref{thm:mixTrans} results in a regret bound in terms of the regret of an arbitrary expert selection sequence $I_1^T\triangleq\{I_t\}_{t=1}^T$.
\begin{definition}\label{def:Es}
	The expert regret of the hyper-expert construction scheme $\Hau$ is given by
	\begin{align}
		E_{S,T}(I_1^T)\triangleq\sum_{s=1}^{S}\sum_{t=T_{s-1}+1}^{T_s} l_t({\bs{\theta}}_{I_t,t})-l_t({\bs{\theta}}_s^*)\nonumber
	\end{align}
	which is the regret of the selections $\{\vt_{I_t}\}_{t=1}^T$ against the competitions $\{\vt_s\}_{s=1}^T$.
\end{definition}

\begin{definition}\label{def:Ws}
	The mixture regret of the transition weighting scheme $\Tau$ for the hyper-expert scheme $\Hau$ is given by
	\begin{align}
		W_{S,T}(I_1^T)\triangleq\Wau_\Tau\left(\{I_t\}_{t=1}^T\right),\nonumber
	\end{align}
	which is the mixture redundancy of the selections $\{\vt_{I_t}\}_{t=1}^T$ in our algorithm.
\end{definition}

\begin{corollary}\label{thm:regret}
	Similarly to \cite{gokcesu2021mixable}, the regret resulting from \autoref{thm:mixTrans} can be divided in two parts as
	\begin{align*}
		R_{S,T}\left(\{{\bs{\theta}}_t\}_{t=1}^T\right)
		\triangleq&\sum_{s=1}^{S}\sum_{t=T_{s-1}+1}^{T_s} l_t({\bs{\theta}}_t)-l_t({\bs{\theta}}_s^*)\\
		\leq& E_{S,T}(I_1^T)+\frac{1}{\alpha}W_{S,T}(I_1^T),
	\end{align*}
	where $E_{S,T}(I_1^T)$ is the expert regret of the hyper-expert construction scheme $\Hau$; and $W_{S,T}(I_1^T)$ is the mixture regret of the transition weighting scheme $\Tau$ in a time horizon $T$ when competing against $S$ time segments.
\end{corollary}

\begin{definition}\label{def:S}
	Let $I_1^T$	be a sequence of experts from $\Hau$ that is able to 'compete', i.e., whenever $I_t\neq I_{t-1}$, $I_t^{th}$ expert starts a new run of the base algorithm. Hence, this sequence of experts $\{I_t\}_{t=1}^T$ collectively imply a sequence of runs of the base algorithm (the structure of which depends on $\Hau$), where during each individual run, the competition $\vt_{s}^*$ stays the same. 
	Let $S_T$ be the number of segments in $\{I_t\}_{t=1}^T$ such that
	\begin{align}
		\mathcal{S}(\{t_s\}_{s=1}^S)\triangleq S_T= 1+\sum_{t=2}^{T}\mathbbm{1}_{I_t\neq I_{t-1}},\nonumber
	\end{align}
	where $\mathbbm{1}_x$ is the identity operator. We point out that $S_T\leq o(T)$ for viable learning, where $o(\cdot)$ is the Little-O notation, because, otherwise, the base algorithm cannot sufficiently learn the optimal parameter in its segment. 
\end{definition}

\begin{definition}\label{def:approx}
	We use the following expression
	\begin{align*}
		F(T)=\hat{O}(G(T)),
	\end{align*}
	to denote an asymptotically near upper bound if $F(T)\leq K (G(T))^{1+\epsilon}$, as $T\rightarrow\infty$ for every finite $\epsilon>0$, i.e., the fraction of $F(T)$ over $G(T)$ is asymptotically sub-polynomial in $G(T)$. Although a function $F(\cdot)$ may not necessarily be upper bounded by an order of $G(T)$, i.e., not $O(G(T))$, it may be $\hat{O}(G(T))$, since it is a looser bound.
\end{definition}

\section{Algorithm Design}\label{sec:des}
There are two components that needs to designed in the algorithmic framework, which are the hyper-expert scheme and the weighting scheme. The weighting scheme determines the switches between the specific hyper-experts, which in turn, determines how the base algorithm is utilized. Unlike \cite{gokcesu2021mixable}, we start with a generic design of the hyper-experts.
\subsection{Hyper Expert Scheme Design}
We start the design by creating a number of hyper-experts which run the base algorithm in given time intervals such that
\begin{align}
	\Hau_{\text{D}}: H(i)=\{p_i,s_i,t_i,r_i\}, &&\forall i\label{Hau},
\end{align}
where the hyper-expert parameters
\begin{align*}
	p_i\in\{1,2,3,\ldots\},
\end{align*}
is the period (i.e., how long does the base algorithm run before restarting) of the $i^{th}$ expert, and
\begin{align*}
	s_{i}\in\{1,2,3,\ldots\},
\end{align*}
is the initial start time (i.e., when we start the run of the base algorithm for the first time) of the $i^{th}$ expert, 
\begin{align}
	t_i\triangleq rem\left((t-s_i),p_i\right)+1
\end{align}
is the current runtime of the base algorithm at time $t$ (i.e., the time since the last start of the base algorithm) for the expert $i$, and
\begin{align}
	r_i\triangleq t-t_i+p_i+1
\end{align} 
is the next reset at time $t$ (i.e., the next time instance the base algorithm resets) for the expert $i$.

\subsection{Wighting Scheme Design}
To mix these hyper-experts, we need a weighting scheme as in \autoref{def:Tau}. The problem setting is not as generic as \cite{gokcesu2021mixable} and the base algorithm does not need to know its total runtime at its start. We use the following weighting scheme
\begin{align}
	\Tau_{\text{D}}:\enspace\tau_t(i,j)=
	\begin{dcases}
		\begin{aligned}
			\frac{1}{t_i},& &&\text{if } t_j=1 \text{ and } j=J_t\\
			\frac{t_i-1}{t_i},& &&\text{if } i=j, t_i\neq1\\
			0,& && \text{ otherwise }
		\end{aligned}
	\end{dcases},\label{quad.n}
\end{align}
where $J_t$ is the index of the expert with the greatest period that resets at time $t$ (tie breaker does not matter), i.e.,
\begin{align}
	J_t\triangleq\argmax_{i:t=r_i}p_i.
\end{align}
Note that the weighting scheme $\Tau_{D}$ satisfies \autoref{def:Tau} and is a valid weighting scheme, since it either stays on the same expert or switches to a single expert at any time $t$.

\subsection{Complexity}
The complexity is directly related to the number of parallel running hyper-experts in accordance with the hyper-expert scheme $\Hau_D$. Thus, the computational complexity at time $t$ is given by the cardinality of the set $C_t\triangleq\{H(i)\in\Hau_{\text{D}}:s_i\leq t\}$. Since this is a growing set, the per time complexity for a time horizon $T$ is bounded by $C_T\triangleq\{H(i)\in\Hau_{\text{D}}:s_i\leq T\}|$.

\subsection{Regret Analysis}
For a given $S$ segment competition, where we compete against parameters $\{\vt_s^*\}_{s=1}^S$ with time lengths $\{t_s\}_{s=1}^S$, let there be a hyper-expert selection sequence $\{I_1^T\}$ that implies $S_T$ restarts of the base algorithm.

\begin{lemma}\label{thm:expertR}
	We have the following expert regret for the scheme $\Hau_D$
	\begin{align*}
		E_{S,T}(I_1^T)=O\left(S_T\log\left(\frac{T}{S_T}\right)\right).
	\end{align*}
	\begin{proof}
		The base algorithm can restart either when we are at the same hyper-expert or when we switch to a different hyper-expert (from the weighting scheme $\Tau_D$, we switch to an expert only when it restarts the base algorithm). Let $\tau_s$ be the total runtime of the base algorithm whenever we go to a new start of the base algorithm. Then, the regret is 
		\begin{align}
			E_{S,T}(\Hau_D)=\sum_{s=1}^{S_T}O(\log(\tau_s)),
		\end{align}
		from the base algorithm regret in \autoref{ass:RB}. Since
		\begin{align}
			\sum_{s=1}^{S_T}t_s=T,
		\end{align}
		we have from concavity of the logarithm
		\begin{align}
			E_{S,T}(\Hau_D)=O\left(S_T\log\left(\frac{T}{S_T}\right)\right),
		\end{align}
		which concludes the proof.
	\end{proof}
\end{lemma}

\begin{lemma}\label{thm:mixtureR}
	We have the following mixture regret for the scheme $\Tau_D$
	\begin{align*}
		W_{S,T}(\Tau_D)=O\left(S_T\log\left(\frac{T}{S_T}\right)\right).
	\end{align*}
	\begin{proof}
		From $\Tau_D$, we stay at the same expert for a runtime $t_i$ of the base algorithm and incur $\log(t_i)$ redundancy during our stay. This is the total redundancy if we stay at the same expert. However, if we switch, we incur an additional $\log(t_i+1)$ redundancy. Thus, we have
		\begin{align}
			W_{S,T}(\Tau_D)\leq \sum_{s=1}^{S_T}2\log(t_s+1),
		\end{align}
		which, from the concavity of the logarithm, gives
		\begin{align}
			W_{S,T}(\Tau_D)=O\left(S_T\log\left(\frac{T}{S_T}\right)\right),
		\end{align}
		which concludes the proof.
	\end{proof}
\end{lemma}

\begin{theorem}
 	The regret bound for using $\Hau_D$ and $\Tau_D$ with \autoref{alg:base} is given by
 	\begin{align*}
 		R_{S,T}\left(\{\hat{\bs{\theta}}_t\}_{t=1}^T\right)=O\left(S_T\log\left(\frac{T}{S_T}\right)\right),
 	\end{align*}
 	for finite mixability $\alpha$, where $S_T$ is the total number of times the base algorithm resets.
 	\begin{proof}
 		The proof is straightforward from the combination of \autoref{thm:expertR} and \autoref{thm:mixtureR} in accordance with \autoref{thm:regret}.
 	\end{proof}
\end{theorem}

\section{Quadratic Complexity Algorithm with Optimal~Logarithmic~Per~Switch~Regret}\label{sec:lin}

For the optimal achievable regret, we need to have a base algorithm starting at each time $t$, since we do not know the time indices the competition changes. To do this, we create an expert at each time $t$ that never resets and continues to sequentially run the base algorithm similarly with \cite{chernov2009,hazan2009}.

We start the design by creating a number of hyper-experts which run the base algorithm in given time intervals such that
\begin{align}
	\Hau_{\text{lin}}: H(i)=\{s_i,p_i,t_i,r_i\}, &&\forall i\label{HauLin},
\end{align}
where the hyper-expert parameters
\begin{align}
	p_i=\infty,
\end{align}
is the period (i.e., the algorithm never restarts) of the $i^{th}$ expert, and
\begin{align}
	s_{i}\in\{1,2,3,\ldots\},
\end{align}
is the initial start time (i.e., when we start the run of the base algorithm for the first time) of the $i^{th}$ expert, 
\begin{align}
	t_i= t-s_i+1
\end{align}
is the current runtime of the base algorithm at time $t$ (i.e., the time since the last start of the base algorithm) for the expert $i$, and
\begin{align}
	r_i=\begin{cases}
		s_i, &t\leq s_i,\\
		\infty, &\text{otherwise}
	\end{cases}
\end{align} 
is the next reset at time $t$ (i.e., the next time instance the base algorithm resets) for the expert $i$.

For this design of the hyper-expert scheme, we have the following complexity and regret.
\begin{corollary}
	For a time horizon $T$, the cardinality bound of the parallel running experts is
	\begin{align*}
		C_T=T,
	\end{align*}
	which results in a linear per time complexity, i.e., $O(T)$ per time.
	\begin{proof}
		The proof comes from the fact that the set of hyper-experts grows linearly (at each time instance, we create a new hyper-expert).
	\end{proof}
\end{corollary}

\begin{corollary}
	For a time horizon $T$, the number of base algorithm restarts $S_T$ is bounded as
\begin{align*}
	S_T= S,
\end{align*}
	which results in the regret bound
	\begin{align*}
		R_{T,S}=O\left(S\log\left(\frac{T}{S}\right)\right)
	\end{align*}
\begin{proof}
	The proof comes from the fact that at each time $t$, there is only one hyper-expert that starts the base algorithm, which we switch to. Thus, for every change in the competition, we switch to a different hyper-expert.
\end{proof}
\end{corollary}

\begin{remark}
	$\Hau_{lin}$ implies an inefficient linear per time (i.e., $O(T)$) algorithm with the optimal regret per switch $O(\log(T/S))$.
\end{remark}

\section{Linearithmic Complexity Algorithm with Sub-Optimal~Log-Squared~Per~Switch~Regret}\label{sec:log}
For efficiency, the traditional approach is to use the doubling trick \cite{doubling_trick} where we run the algorithm with time lengths that is the double of the previous run (e.g., $2^0, 2^1, 2^2, \ldots$). Hence, the hyper-experts end up running the base algorithm for lengths that are powers of $2$. We efficiently implement our algorithm by creating hyper experts that rerun the base algorithm in specific time intervals. These hyper-experts will be designed such that the $i^{th}$ expert, $i\in\{1,\ldots,N\}$, will run the base algorithm individually in subsequent time segments of length $2^k$ for some $k$ similarly with \cite{gokcesu2020recursive,gokcesu2021mixable}. 

We design the hyper-experts with the parameter $k_i$, which is a parameter such that
\begin{align}
	k_i\in\{2^0,2^1,2^2,2^3,\ldots\},
\end{align}
where the hyper-expert $i$ with $k_i$ start its run at $t=k_i$; and runs the base algorithm for an interval of $k_i$. Thus,
\begin{align}
	\Hau_{\text{log}}: H(i)=\{s_i,p_i,t_i,r_i\}, &&\forall i\label{HauLog},
\end{align}
where the hyper-expert parameters
\begin{align}
	p_i\in\{2^0,2^1,2^2,2^3,\ldots\},
\end{align}
is the period of the $i^{th}$ expert, and
\begin{align}
	s_{i}=p_i,
\end{align}
is the initial start time of the $i^{th}$ expert, 
\begin{align}
	t_i= rem((t-p_i),p_i)+1
\end{align}
is the runtime of the base algorithm at $t$ for the expert $i$, and
\begin{align}
	r_i=t-t_i+p_i+1
\end{align} 
is the next reset at time $t$ (i.e., the next time instance the base algorithm resets) for the expert $i$.

\begin{corollary}
	For a time horizon $T$, the cardinality bound of the parallel running experts is
	\begin{align*}
		C_T=O(\log(T)),
	\end{align*}
	which results in a logarithmic per time complexity.
	\begin{proof}
		The proof comes from the fact that there is only one hyper-expert with the same period and this set grows logarithmically with time.
	\end{proof}
\end{corollary}

\begin{corollary}
	For a time horizon $T$, the number of base algorithm starts $S_T$ is bounded as
	\begin{align*}
		S_T= S\log(T/S),
	\end{align*}
	which results in the regret bound
	\begin{align*}
		R_{T,S}=O\left(S\log^2\left({T}/{S}\right)\right)
	\end{align*}
	\begin{proof}
		From the construction of $\Hau_{log}$, we see that if we switch to a period $p_i$ at time $t$ for some competition $\vt_s^*$, we can only switch to a bigger period at $t+p_i$. This results in the worst case scenario of $\log(t_s)$ restarts for the competition segment length $t_s$, which concludes the proof.
	\end{proof}
\end{corollary}

\begin{remark}
	$\Hau_{log}$ implies an efficient logarithmic per time (i.e., $O(\log(T))$) algorithm with sub-optimal regret per switch $O(\log^2(T/S))$.
\end{remark}

\section{Near-Linear Complexity Algorithm with Near-Optimal~Near-Logarithmic~Per~Switch~Regret}\label{sec:opt}
In this section, we combine the better properties of the designs in \autoref{sec:lin} and \autoref{sec:log}. We leave the period structure generic, which is not necessarily an exponentially growing one as in \autoref{sec:log}
\begin{align}
	p_i\in\{f_1,f_2,\ldots\},
\end{align}
where $f_1=1$.
For $n\geq 2$, let $f_n=\alpha_nf_{n-1}+\beta_n$, where $\alpha_n$ is a natural number and $0\leq\beta_n\leq f_{n-1}$.
We design the hyper experts such that for every period $f_n$, the start times will range from $\{\beta_n+f_{n-1},\beta_n+2f_{n-1},\ldots,\beta_n+\alpha_nf_{n-1}\}$.
Thus,
\begin{align}
	\Hau_{\text{sub}}: H(i)=\{s_i,p_i,t_i,r_i\}, &&\forall i\label{HauSub},
\end{align}
where the hyper-expert parameters
\begin{align}
	p_i=f_n\in\{f_1,f_2,\ldots\},
\end{align}
for some $n$, is the period of the $i^{th}$ expert, and
\begin{align}
	s_{i}\in\{\beta_n+f_{n-1},\beta_n+2f_{n-1},\ldots,\beta_n+\alpha_nf_{n-1}\},
\end{align}
is the initial start time of the $i^{th}$ expert, 
\begin{align}
	t_i= rem((t-s_i),p_i)+1
\end{align}
is the runtime of the base algorithm at $t$ for the expert $i$, and
\begin{align}
	r_i=t-t_i+p_i+1
\end{align} 
is the next reset at time $t$ for the expert $i$.

\begin{remark}
	When $\alpha_n=1$ for all $n$ and $\beta_n=f_{n-1}$, we have the structure of \autoref{sec:log} as in \cite{gokcesu2020recursive,gokcesu2021mixable}.
\end{remark}

\begin{definition}\label{def:nt}
	We define the following metric which will be useful in both the complexity and the regret analysis. Let
	\begin{align*}
		n_t=\max_{n:f_n<t}n,
	\end{align*}
	which is the index of the maximum period that is less than the time length $t$.
\end{definition}

\begin{theorem}\label{thm:CT}
	For a time horizon $T$, the cardinality bound of the parallel running experts is
	\begin{align*}
		C_T\leq 1+n_T\max_{n\in\{2,\ldots,n_T+1\}}\alpha_n,
	\end{align*}
	where $n_T$ is as in \autoref{def:nt}.
	\begin{proof}
		We observe that for an expert with $f_n$ period to start for the first time, we need to have passed the time mark $f_{n-1}$. For a time horizon $T$, we have from \autoref{def:nt}
		\begin{align}
			n_T=\max_{n:f_{n}<T}n
		\end{align}
	Then, complexity is
	\begin{align}
		C_T\leq&1+\sum_{n=1}^{n_T}\alpha_{n+1},\\
		&\leq 1+n_T\max_{n\in\{2,\ldots,n_T+1\}}\alpha_n,
	\end{align}
	where the additive $1$ comes from the single hyper-expert with the period $f_1=1$, which completes the proof.
	\end{proof}
\end{theorem}

\begin{theorem}\label{thm:ST}
	For a time horizon $T$, $\Hau_{sub}$ results in the regret
	\begin{align*}
		R_{T,S}=O\left(\left(S+\sum_{s=1}^Sn_{t_s}\right)\log\left(\frac{T}{S+\sum_{s=1}^Sn_{t_s}}\right)\right),
	\end{align*}
	where $n_{t_s}$ is as in \autoref{def:nt}.
\begin{proof}
	For any competition segment $t_s$, let
	\begin{align}
		n_{t_s}=\max_{n:f_n<t_s}n.
	\end{align}
	Whenever we switch to an expert with $f_{n_{t_s}+1}$, we can cover until the end of the competition. Moreover, we do not need to switch to the same period twice. Thus, total number of switches is bounded by $n_{t_s}+1$ and $S_T\leq S+\sum_{s=1}^Sn_{t_s}$.
\end{proof}
\end{theorem}

We design the periods $f_n$ for $n\geq2$ as the following
\begin{align}
	f_n=\lfloor \exp(a\exp(b\log^c(n))) \rfloor,\label{fn}
\end{align}
where $a>0$, $b>0$, $c>1$. 
From \autoref{def:nt}, we have
		\begin{align}
			n_{t_s}\leq&\exp((\log(\log(t_s+1)/a)/b)^{1/c}),\label{nts}
		\end{align}
		which is sub-logarithmic for finite $c>1$.

\begin{corollary}
	For a time horizon $T$, the cardinality bound of the parallel running experts is asymptotically
	\begin{align*}
		C_T=o(T^\epsilon),
	\end{align*}
	for any finite $\epsilon>0$, i.e., sub-polynomial.
	\begin{proof}
		For $c>1$, we have
		\begin{align}
			\alpha_{n_T}
			=&O\left( T^{\frac{bc\log^{c-1}(n_T)}{n_T}}\right),
		\end{align}
		from asymptotic convexity of $\exp(b\log^c(n))$. The result comes from \autoref{thm:CT} and the fact that $n_T$ is sub-logarithmic and divergent. 
	\end{proof}
\end{corollary}

\begin{corollary}
	Using the periods in \eqref{fn} results in the regret
	\begin{align*}
		R_{T,S}=\hat{O}\left(S\log\left(T/S\right)\right),
	\end{align*}
	since $S_T$ is sub-logarithmic per $S$.
	\begin{proof}
		Using \eqref{nts} and \autoref{thm:ST}, we have
		\begin{align}
			S_T= O(S\exp((\log(\log(T/S+1)/a)/b)^{1/c})),
		\end{align}
		which results in a near-logarithmic regret per switch. 
	\end{proof}
\end{corollary}

\begin{remark}
	$\Hau_{sub}$ implies an efficient near-linear time algorithm with near-optimal near-logarithmic regret per switch.
\end{remark}

\section{Conclusion}\label{sec:con}
We studied the online optimization of mixable loss functions in a dynamic environment, where there exists an algorithm that achieves $O(\log(T))$ regret against a fixed estimator. We proposed an online mixture framework that uses these static solvers as its base algorithm. Against a competition with $S$ segments, we showed that, with the suitable selection of hyper-expert creations and weighting strategies, we can achieve $O(S\log(T/S))$ and $O(S\log^2(T/S))$ regret in $O(T)$ and $O(\log(T))$ computational complexity per round, respectively. Furthermore, we showed that it is also possible to achieve asymptotically ${O}(S\log^{1+\epsilon}(T/S))$ regret for any finite $\epsilon>0$ (near-logarithmic, thus near-optimal) with $O(T^{1+\delta})$ computational complexity for any finite $\delta>0$ (near-linear).

\bibliographystyle{IEEEtran}
\bibliography{double_bib}

\end{document}